%% file: main.tex
\documentclass{article}
\usepackage{spconf,amsmath,graphicx}
\usepackage{mathtools}
\usepackage[dvipsnames]{xcolor}
\usepackage{pifont}
\usepackage{soul}
\usepackage{multirow}
\usepackage{verbatim}
\usepackage{pgfplots}
\pgfplotsset{compat=1.17}
\renewcommand{\paragraph}[1]{\noindent\textbf{#1}\quad}
\usepackage[color=true, onerror=nice]{yagusylo}
\defdingname[fourier][global]{116}{rhand}[RubineRed]
\usepackage{dirtytalk}
\usepackage{booktabs}
\usepackage[pagebackref=true,breaklinks=true,colorlinks,bookmarks=false]{hyperref}
\newcommand{\yang}[1]{}  

\usepackage[textsize=footnotesize]{todonotes}

\usepackage{subcaption}
\usepackage{pgfplots}
\usepackage{interval}

\newcommand{\datasetName}{QuerYD}

\let\OLDsection\section
\renewcommand\section[1]{\OLDsection{#1}\vspace{-3pt}}

\let\OLDthebibliography\thebibliography
\renewcommand\thebibliography[1]{
  \OLDthebibliography{#1}
  \small
  \setlength{\parskip}{0pt}
  \setlength{\itemsep}{1pt plus 0.3ex}
}

\title{QuerYD: a
video dataset with high-quality text and audio narrations}
\name{Andreea-Maria Oncescu\quad Jo\~ao F. Henriques \quad Yang Liu \quad Andrew Zisserman \quad Samuel Albanie}

\address{Visual Geometry Group, University of Oxford, UK\\
\url{http://www.robots.ox.ac.uk/~vgg/data/queryd}
}

\begin{document}
%
\maketitle
\begin{abstract}
We introduce \datasetName{}, a new large-scale dataset for retrieval and event localisation in video.
A unique feature of our dataset is the availability of two audio tracks for each video: the original audio, and a high-quality spoken description of the visual content.
The dataset is based on YouDescribe \cite{youdescribe},
a volunteer project that assists visually-impaired people by attaching voiced narrations to existing YouTube videos.
This ever-growing collection of videos
contains highly detailed, temporally aligned
audio and text annotations.
The content descriptions are more relevant than dialogue, and more detailed than previous description attempts, which can be observed to contain many superficial or uninformative descriptions.
To demonstrate the utility of the \datasetName{} dataset, we show that it can be used to train and benchmark strong models for retrieval and event localisation.
Data, code and models are made publicly available, and we hope that \datasetName{} inspires further research on video understanding with written and spoken natural language. 
\end{abstract}
\begin{keywords}
Audio description, retrieval
\end{keywords}
\medskip
\input{intro.tex}

\input{related.tex}

\input{dataset.tex}
\input{experiments.tex}
\input{conclusion.tex}
\input{ack}



\bibliographystyle{IEEEbib}
\bibliography{refs}

\end{document}

%% file: intro.tex
\section{Introduction} \label{sec:intro}

The development of new datasets has been instrumental in the immense progress of modern machine learning, working hand-in-hand with methodological improvements.
New, larger datasets allow incremental gains in performance with no change in methodology \cite{Hammar_2018}, and prevent progress stalling from over-fitting to saturated benchmarks \cite{recht2018cifar}.
In this work, we propose a dataset that supports investigations into the relationship between video and natural language.

For this task, obtaining rich and diverse training data is essential, an endeavour that is made difficult by the fact that ``describing a video'' is a relatively under-constrained objective.
Dataset designers that rely on crowdsourced annotations, usually through paid micro-work platforms such as Amazon Mechanical Turk, must specify elaborate guidelines and multi-stage verification mechanisms to attempt to remove low-effort solutions, given the monetary incentive \cite{vatex}.

We propose a new dataset for retrieval and localisation, sourced from the \emph{YouDescribe} community\footnote{\url{http://youdescribe.ski.org}} that contributes audio descriptions for YouTube videos.
The videos cover diverse domains and the descriptions are precisely localised in time. They exhibit richer vocabulary than existing publicly available text-video description datasets (sec.~\ref{sec:video_understanding_task}).

The aim of the annotators is to \emph{communicate} the contents of videos to visually-impaired people, resulting in diverse and high quality audio descriptions.
Annotators add spoken narrations to videos from YouTube, subject to the time-constraint of describing the action in near real-time.  
%
\begin{figure*}
    \centering
    \vspace{-0.5cm}
    \includegraphics[width=0.9\textwidth]{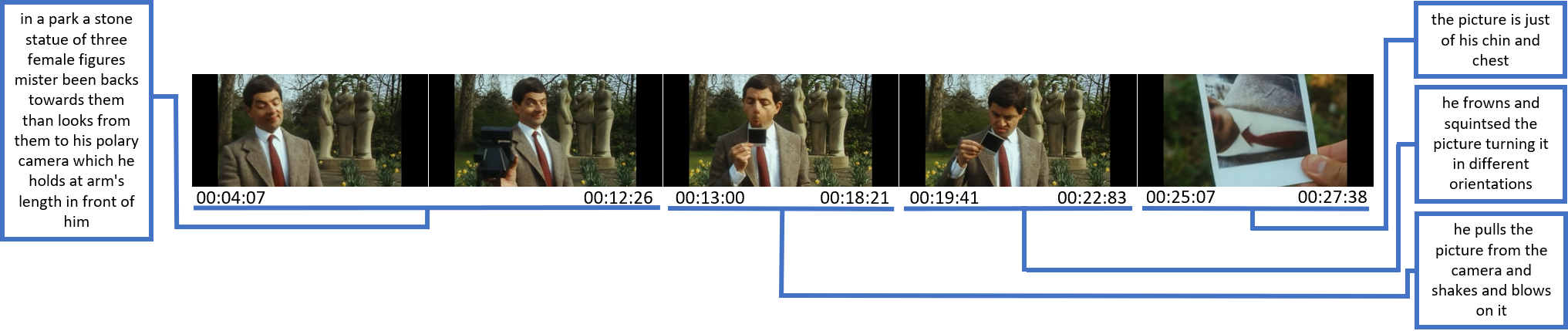}
    \includegraphics[width=0.9\textwidth]{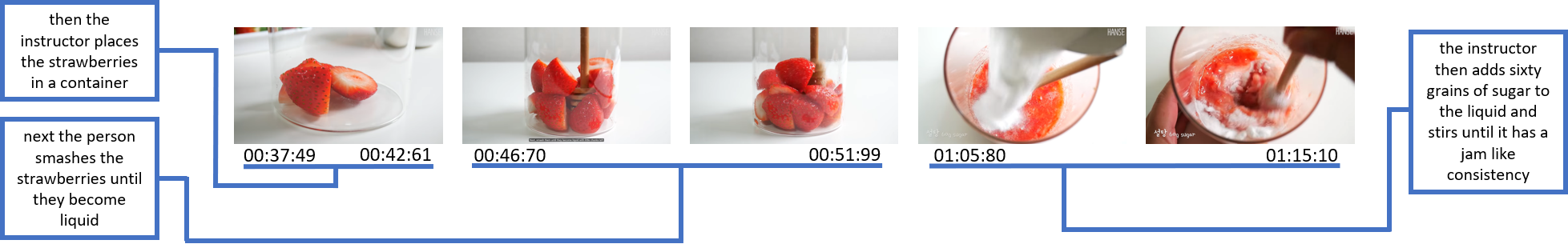}
    \caption{Qualitative examples from the \datasetName{} dataset. Audio descriptions for video segments were provided and transcribed by voluntary annotators. Some frames are individually described while other annotations take place over a sequence of frames. The time localisation of the audio descriptions is presented in the format \emph{minutes:seconds:miliseconds}. \yang{Do we need an examples that contains both speech and audio narrations, to illustrate they might be different/complementary? If we want to highlight this is a new modality}}
    \label{fig:qual-examples-time}
    \vspace{-0.3cm}
\end{figure*}
The proposed dataset expands previous datasets across four axes: (1) \textit{Modality.} In addition to the original  audio track of each video, \datasetName{} contains  a separate audio track containing spoken narrations describing the action. This modality is highly complementary to the standard audio, the content of which can be unrelated to the visual content (e.g.\ background music, character dialogue about off-screen events, and aspects of the action that do not produce sound). In contrast, spoken narrations have a one-to-one relation to the video's content, encoded as audio. (2) \textit{Quantity.} \datasetName{} is large-scale, containing over 200 hours of video and 74 hours of audio descriptions.
It also contains a higher density of descriptions than other datasets (as measured by words-per-second). (3) \textit{Quality.} Since the descriptions in \datasetName{} are created by volunteers who aim for high-quality descriptions of videos for the visually-impaired, and narrations are rated by other users, there is an added incentive for quality when compared to micro-work platforms. We demonstrate this difference empirically, observing a larger vocabulary size, larger number of sentences per video, and sentence lengths, when compared to other datasets (which generally contain text captions instead of audio narrations). (4) \textit{Scalability.} \datasetName{} is not a static dataset, since it is based on an ever-growing collection of audio descriptions, continually evolving with new narrations added every day. By periodically updating this dataset,
we will empower future researchers to obtain more up-to-date snapshots of the audio descriptions data, ensuring that it stays relevant as the data demand grows.

In summary, our contributions are: (1) we propose \datasetName{}, a new dataset for video retrieval and localisation; (2) we provide an analysis of the data, demonstrating that it possesses a richer diversity of text descriptions than prior work; (3) we provide baseline performances with existing state-of-the-art models for text-video understanding tasks. 



%% file: related.tex
\section{Related Work} \label{ref:related}

Researchers in the computer vision and natural language processing communities have been striving to bridge the gap between videos and natural language. We have seen significant progress in many tasks, such as text-to-video retrieval, text-based action or event localisation, and video captioning. This success has resulted both from advances in deep learning as well as from the availability of video description datasets.

\paragraph{Video-Text Retrieval Benchmarks.} In the past, video-text retrieval datasets have addressed controlled settings \cite{barbu2012video, kojima2002natural}, specific domains such as cooking \cite{rohrbach2014coherent, das2013thousand}, and acted and edited material in movies~\cite{rohrbach2015dataset,yao2015describing,bain2020condensed}. Some new video datasets \cite{xu2016msr, chen2011collecting, anne2017localizing,krishna2017dense,Wang_2019_ICCV} have been collected from YouTube and Flicker, which are open-domain and realistic. 
However, in each case, obtaining text annotations for these datasets is time-consuming and expensive and therefore difficult to do at large scale.
Some recent works obtained text annotations by automatically transcribing narrations instead, and collected relatively large-scale datasets, such as CrossTask\cite{zhukov2019cross}, YouCook2 \cite{ZhXuCoCVPR18}, HowTo100M \cite{miech2019howto100m}.
However, these datasets are all sourced from instructional videos related to pre-defined tasks.
Moreover, collecting annotations from narration introduces some incoherence (noise) in the video-text pairs, due to the narrator talking about things unrelated to the video, or describing something before or after it happens.

In this work, we explore a different source for text annotations -- Audio Description (AD). AD provides descriptions of the visual content in a video which allows visually-impaired people to understand the video. Compared to manual annotations and narration transcriptions from instructional videos, ADs describe precisely what is shown on the screen 
and are temporally aligned to the video. Related to our work, LSMDC \cite{rohrbach2015dataset} also considers AD as a source of supervision.  However, as opposed to our work, LSMDC focuses on movies and short clips (3s on average).  Also of relevance is the Epic Kitchens dataset~\cite{epicKitchens}, which employs narration to provide descriptions, but differs from our approach through its focus on egocentric videos of kitchen-based activities.
Our dataset 
instead aims to cover a more general range of domains. 

\paragraph{Event Localisation Benchmarks.} Event localisation is a task that aims to retrieve a specific temporal segment from a video given a natural language text description. It requires the event localisation methods to determine not only what occurs in a video but also when. Some datasets have been collected for this task in the past few years, including Charades-STA \cite{gao2017tall}, DiDemo \cite{anne2017localizing}, and the Activity-Net \cite{krishna2017dense} dataset. However, the average length of the video in these datasets is less than 180s and the number of events per video is less than 4, which makes the event localisation less challenging. In contrast, our \datasetName{} dataset consists of longer videos and more distinct moments from each unedited video footage paired with descriptions that can uniquely localise the moment. It is also worth noting that instead of using manual annotations which can suffer from ambiguity in the event definition and disagreement of the start and end times, AD provide more fine-grained and precise temporal segment annotations.

%% file: dataset.tex
\section{The \datasetName{} Dataset} \label{sec:method}

In this section, we first give a high-level overview of the \datasetName{} dataset.  Afterwards, we provide qualitative examples and quantitative analyses of the videos and descriptions that comprise the dataset, compared to others in the literature.

\begin{figure}
    \centering
    \includegraphics[width=0.70\columnwidth]{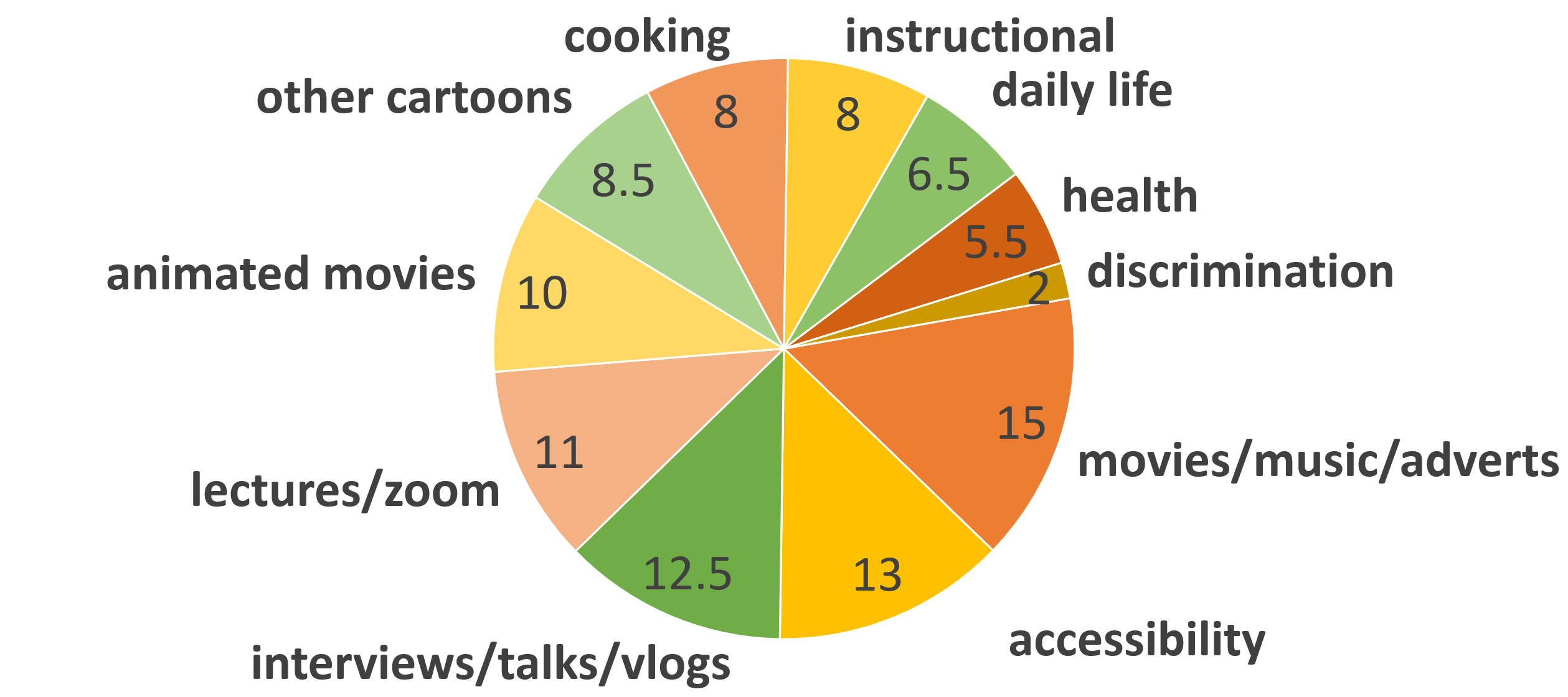}
    \caption{Detailed classification of the \datasetName{} video content by category, and associated proportion (\%).}
    \label{fig:data_split}
\end{figure}

\paragraph{Dataset overview.} The \datasetName{} dataset comprises 207 hours of video accompanied by 74 hours of Audio Description (AD) transcriptions, resulting in 31,441 descriptions. The videos, which are sourced from YouTube, cover a diverse range of visual content shown in Fig.~\ref{fig:data_split}.  Of the total transcriptions, 13,019 are localised within video content (with precise start and end times), and the remaining 18,422 descriptions are \textit{coarsely} localised (they are assigned a single time in a video, but their temporal extent is not annotated).
The training, validation and test partitions of the dataset, obtained by uniform sampling of videos, are shown in Tab.~\ref{table:partitions}.

\begin{table}
\scriptsize
\setlength{\tabcolsep}{8pt}
    \centering
    \begin{tabular}{l | c c c}
    \toprule
        Partition  & Train & Validation & Test \\
        \midrule
        Untrimmed videos & 1,815 & 388 & 390 \\
        Total descriptions & 22,008 & 4,716 & 4,717  \\
        Localised descriptions & 9,113 & 1,952 & 1,954 \\
    \bottomrule
    \end{tabular}
    \caption{The partitions of the proposed \datasetName{} dataset.}
    \label{table:partitions}
\end{table}

\begin{figure}
\centering
\begin{minipage}{\columnwidth}
  \centering
    \includegraphics[width=0.5\textwidth]{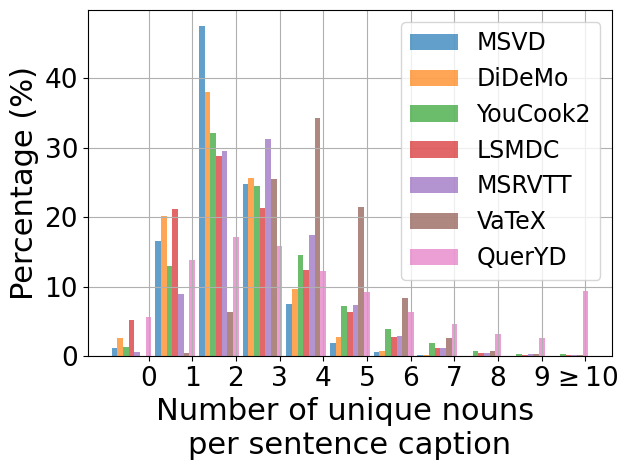}%
    \includegraphics[width=0.5\textwidth]{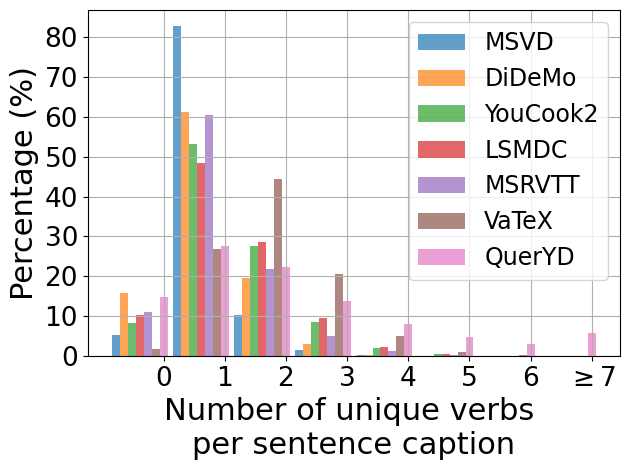}%
    \captionof{figure}{A comparison of the linguistic complexity and diversity of \datasetName{} with popular text-video datasets.
    Note that \datasetName{} has a higher proportion of unique nouns and verbs.}
    \label{fig:vocab-comparison}
\end{minipage}
\end{figure}

\paragraph{Dataset analysis.} We analysed several aspects of the \datasetName{} dataset: the diversity of vocabulary (Fig.~\ref{fig:vocab-comparison}), linguistic complexity (Tab.~\ref{table-stats}), as well as the video duration and the distribution of localised segments (Tab.~\ref{table-stats}).

We compare the \datasetName{} dataset with other datasets containing annotated videos such as LSMDC\cite{rohrbach2015dataset},  VaTeX \cite{Wang_2019_ICCV}, YouCook2 \cite{ZhXuCoCVPR18}. Because the \datasetName{} videos are chosen by voluntary annotators from YouTube, the topics vary greatly with a vocabulary more varied than that of more specific datasets, as can be seen in Fig. \ref{fig:vocab-comparison} and Table \ref{table-stats}. The vocabulary size is calculated by first assigning words to their corresponding part-of-speech and then lemmatising the tokens using the Spacy library~\cite{spacy}. As well as having more varied video topics, the \datasetName{} dataset has a wider range of video lengths as demonstrated in Tab.~\ref{table-stats}, with 9 videos longer than 40 minutes. The average video time is 278 seconds for untrimmed video, while the average of the video segments is of 7.72 seconds. This results in better temporal localisation since the audio description provided describes only the current scene.

\paragraph{Qualitative examples.} \label{sec:qualitative} In this section, we explore some qualitative examples from \datasetName{}. Examples of videos and captions are given in Fig.~\ref{fig:qual-examples-time} together with the timestamps of the corresponding audio descriptions. In the top picture there are four rich descriptions covering 27 seconds of one video. These descriptions have a high content of verbs and nouns, reflecting the statistics in Fig.~\ref{fig:vocab-comparison}.
The bottom image is another example of detailed descriptions of a short video with extremely fine-grained localisation.


\paragraph{Descriptions localisation.} To evaluate how well localised audio descriptions are, 95 audio descriptions were randomly selected and start and end times were carefully annotated. Based on these values, the mean IOU metric for time intervals was found to be 0.78. When calculating the time difference in seconds between description's start and end times and the corresponding annotated values, a mean difference of three seconds was obtained.

\paragraph{Data collection.} \datasetName{} is gathered from user-contributed descriptions provided by the \emph{YouDescribe} community, who contribute audio descriptions to videos hosted on YouTube to assist the visually-impaired~\cite{youdescribe}. A portion of these audio descriptions is further accompanied by user-provided transcriptions.  To handle cases in which such a transcription is not provided, we use the Google Speech-to-Text API~\cite{google-speech} to transcribe audio descriptions.
Unlike speech transcription from speech \say{in the wild}, audio descriptions are recorded in a clean environment and contain only the voice of the speaker, leading to high quality transcriptions.
The \emph{YouDescribe} privacy policy~\cite{youdescribe} ensures that all captions are public with the consent of annotators. Lastly, we also group the localised descriptions with their corresponding clips for the localisation task and eliminate empty audio descriptions.  


\input{tables/stat}

%% file: tables/stat.tex
\begin{table}[!t]
\scriptsize
\centering
\setlength{\tabcolsep}{4pt}
\resizebox{\linewidth}{!}{
\begin{tabular}{c|c|c|c|c|c|c|cccccc}
\hline
\hline  
\multirow{2}{*}{Dataset} &\multirow{2}{*}{\#Videos} & Video &\multirow{2}{*}{\#Clips}& Clip & \#Sentence & Sent Length  & \multicolumn{5}{c}{Vocabulary}\\
 & &Length &&Length& per video& per clip &Total&Noun&Verb&Adj&Adverb\\
\hline
DiDeMo & 9605 & 25s &37185 & 6.3s & $5.86\pm0.34$& $7.50\pm 3.23$ & 7865 & 3475 & 1316 & 841 & 339 \\
ACT & 20,000 & 180s &54926&36.18s & $3.56\pm1.67$ & $13.58\pm6.44$ & 12413  & 5218 & 2162 & 1590 & 534\\
QuerYD & 2593 & 278s & 13019 & 7.72s & $12.25\pm15.27$ & $19.91 \pm{22.89}$ & 28515  & 8825 & 3551 & 3128 & 907\\
\hline
\hline
\end{tabular}
}
\caption{Comparison of the event localisation datasets. \yang{can we double check the variance of our dataset? quite big :)}}
\label{table-stats}
\end{table}

%% file: experiments.tex
\section{Video Understanding Tasks} \label{sec:video_understanding_task}

In this section we demonstrate the application of \datasetName{} to two video understanding tasks: paragraph-level video retrieval
and clip localisation.
We consider three models:

\noindent (1) The \textit{E2EWS} (End-to-end Weakly Supervised) model proposed by~\cite{miech2019end} is a cross-modal retrieval model that is trained using weak supervision from a large-scale corpus (100 million) of instructional videos (using speech content as the supervisory signal). The model employs a S3D~\cite{xie2018rethinking} video feature extractor and a lightweight text encoder. We use the video and text encoders without any form of fine-tuning on \datasetName{}, providing a calibration of task difficulty.  \\
(2) The \textit{MoEE} (Mixture of Embedded Experts) model proposed by~\cite{miech2018learning} comprises a multi-modal video model in combination with a system of context gates that learn to fuse together different pretrained \say{experts} (inspired by the classical Mixture of Experts model~\cite{jordan1994hierarchical}) to form a robust cross-modal text-video embedding.  \\
(3) The \textit{CE} (Collaborative Experts) model similarly learns a cross-modal embedding by fusing together a collection of pretrained experts to form a video encoder.  It uses a relation network~\cite{santoro2017simple} sub-architecture to combine together different modalities, and represents the state-of-the-art on several retrieval benchmarks.\yang{Do we have any estimation about how many videos we have speech, if more than half, maybe it is interesting to see whether speech is useful for retrieval for this dataset}
\noindent 
Except where otherwise noted, the MoEE and CE models adopt the same four pretrained experts for scene classification, action recognition, ambient sound classification and image classification described in~\cite{liu2019use} (described in detail in the supplemental material). 

\paragraph{Text-video retrieval.} We first consider the task of paragraph-level video-retrieval with natural language queries (e.g.\ as studied in ~\cite{zhang2018cross,liu2019use}). 
We report standard retrieval evaluation metrics, including median rank (MdR, lower is better), mean rank (MnR, lower is better) and recall at rank K (R@k, higher is better). We report the mean and standard deviation for three runs with different random seeds.
We first compare the three different baseline models on our dataset and then investigate the importance of the use of different modalities.



\noindent \textit{Baselines}. Our results, reported in Tab.~\ref{table:retrieval-baselines}, show that the CE model ~\cite{liu2019use} trained on our \datasetName{} dataset performs best, outperforming the MoEE ~\cite{miech2018learning} and E2EWS ~\cite{miech2019end} models (the latter being trained on a corpus of 100 million instructional videos, but used without any form of finetuning).
Additionally, we propose a baseline for retrieval where the full QuerYD dataset is used as test set. The results when running the E2EWS ~\cite{miech2019end} model without finetuning on the full QuerYD dataset are shown in Table \ref{table:retrieval-mnnet}.

\input{tables/retrieval-baselines}

\input{tables/mnnet_full_retrieval_test_queryd}

\noindent \textit{The importance of different modalities for \datasetName{} retrieval}. In Tab.~\ref{table:ablation-modalities}, we assess the importance of different pretrained experts for the retrieval performance of the CE model ~\cite{liu2019use} on \datasetName{}.
We observe that the addition of each expert improves the performance, although to different extents: both object and action classification experts bring large gains in performance, while the ambient sound expert produces mixed results and has a smaller impact, indicating that this is a weaker cue, but one that still could benefit retrieval.

\input{tables/retrieval-modalities}

\paragraph{Clip localisation.} We next study clip localisation. Corpus-level clip localisation without effective proposals is an extremely challenging task (e.g. \cite{escorcia2019temporal} reports a performance of 0.85\% recall at rank 1 on DiDeMo).  For this reason, in this work we establish a baseline for the proposal oracle setting, which consists of having the temporal segment localisations (ground truth proposals) provided as input to each method.


\noindent \textbf{Localisation baselines}. Our results, reported in Tab.~\ref{table:localization-baselines}, show that the MoEE method performs best, followed closely by CE.
The E2EWS method, which was trained with weak supervision from a much larger set of videos, performs worse. This may be due in part to the weak supervision, which does not enforce the network to be highly sensitive to timing. The model nevertheless establishes a solid baseline for performance without fine-tuning.

\input{tables/localization-baselines}

\begin{figure}
    \centering
    \includegraphics[width=0.45\textwidth]{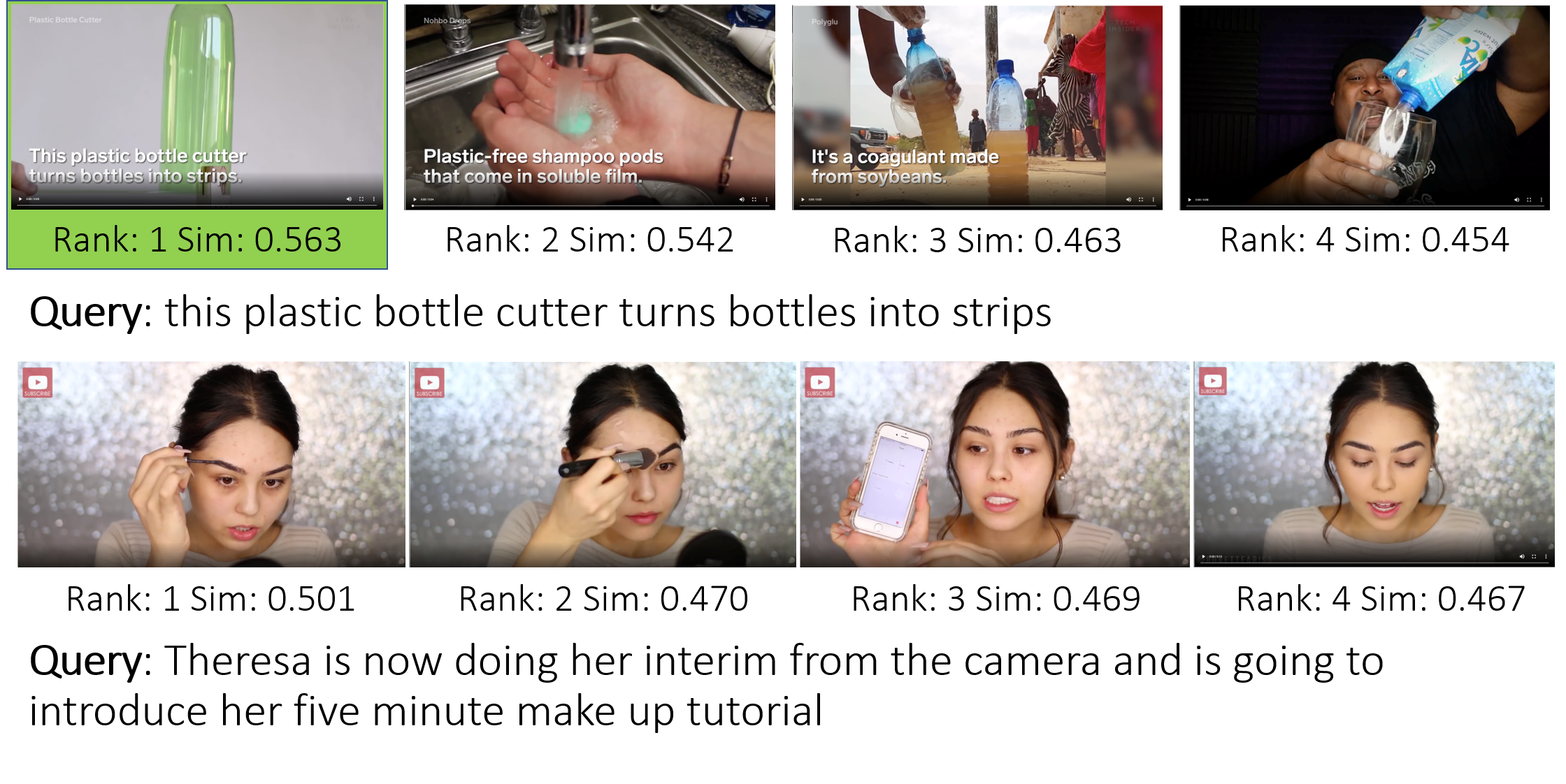}
    \vspace{-1em}
    \caption{Success (top) and failure (bottom) cases of the CE model trained on \datasetName{} and tested on text-video retrieval.}
    \label{fig:qual-examples}
\end{figure}

\noindent \textbf{Failure cases.}
In Fig.~\ref{fig:qual-examples} we show some failure cases of the CE model, which we observed to be the strongest model in the text-video retrieval task.
One of the main failure modes is the ambiguity between segments that can plausibly correspond to the retrieval target, showing that at such a fine level of temporal granularity no model can perform perfectly.






%% file: tables/retrieval-baselines.tex
\begin{table}
\centering 
\scriptsize 
\setlength{\tabcolsep}{6pt}
\resizebox{\columnwidth}{!}{%
\begin{tabular}{l | @{\hskip 0.1cm}c@{\hskip 0.1cm}c@{\hskip 0.1cm}c@{\hskip 0.1cm}c@{\hskip 0.1cm}c|@{\hskip 0.1cm}c @{\hskip 0.1cm}c@{\hskip 0.1cm}c@{\hskip 0.1cm}c@{\hskip 0.1cm}c@{\hskip 0.1cm}c} 
\hline \hline
\multicolumn{1}{c}{} & 
\multicolumn{5}{c}{Text $\implies$ Video} & \multicolumn{5}{c}{Video $\implies$ Text} \\
Method & R$@$1 $\uparrow$ & R$@$5 $\uparrow$ & R$@$10 $\uparrow$ & MdR $\downarrow$ & MnR $\downarrow$ & R$@$1 $\uparrow$ & R$@$5 $\uparrow$ & R$@$10 $\uparrow$ & MdR $\downarrow$ & MnR $\downarrow$ \\ 
\hline 
E2EWS~\cite{miech2019howto100m} & $13.5_{\pm0.0}$&$27.5_{\pm0.0}$&$34.5_{\pm0.0}$&$35.0_{\pm0.0}$ & $72.5_{\pm0.0}$&$ 12.4_{\pm0.0} $&$23.8_{\pm0.0}$&$30.8_{\pm0.0}$&$33.0_{\pm0.0}$&$73.4_{\pm0.0}$ \\
MoEE~\cite{miech2018learning}&$11.6_{\pm1.3}$&$30.2_{\pm3.0}$&$43.2_{\pm3.1}$&$14.2_{\pm1.6}$&$42.7_{\pm2.6}$ & $13.0_{\pm3.1}$&$30.9_{\pm2.0}$&$43.0_{\pm2.8}$&$14.5_{\pm1.8}$&$42.6_{\pm1.5}$\\
CE~\cite{liu2019use} & $13.9_{\pm0.8}$&$37.6_{\pm1.2}$&$48.3_{\pm1.4}$&$11.3_{\pm0.6}$&$35.1_{\pm1.6}$ & $13.7_{\pm0.7}$&$35.2_{\pm2.7}$&$46.9_{\pm3.2}$&$12.3_{\pm1.5}$&$35.8_{\pm2.4}$ \\
\hline \hline

\end{tabular}%
}
\caption{Comparison of text-video retrieval methods trained with paragraph-level information on the \datasetName{} dataset.
} 
\label{table:retrieval-baselines} 
\end{table}

%% file: tables/mnnet_full_retrieval_test_queryd.tex
\begin{table}
\centering 
\scriptsize 
\setlength{\tabcolsep}{6pt}
\resizebox{\columnwidth}{!}{%
\begin{tabular}{l | @{\hskip 0.1cm}c@{\hskip 0.1cm}c@{\hskip 0.1cm}c@{\hskip 0.1cm}c@{\hskip 0.1cm}c|@{\hskip 0.1cm}c @{\hskip 0.1cm}c@{\hskip 0.1cm}c@{\hskip 0.1cm}c@{\hskip 0.1cm}c@{\hskip 0.1cm}c} 
\hline \hline
\multicolumn{1}{c}{} & 
\multicolumn{5}{c}{Text $\implies$ Video} & \multicolumn{5}{c}{Video $\implies$ Text} \\
Method & R$@$1 $\uparrow$ & R$@$5 $\uparrow$ & R$@$10 $\uparrow$ & MdR $\downarrow$ & MnR $\downarrow$ & R$@$1 $\uparrow$ & R$@$5 $\uparrow$ & R$@$10 $\uparrow$ & MdR $\downarrow$ & MnR $\downarrow$ \\ 
\hline 
E2EWS~\cite{miech2019howto100m} & $3.4_{\pm0.0}$&$9.8_{\pm0.0}$&$14.4_{\pm0.0}$&$274.0_{\pm0.0}$ & $523.8_{\pm0.0}$&$ 4.2_{\pm0.0} $&$10.4_{\pm0.0}$&$13.9_{\pm0.0}$&$268.5_{\pm0.0}$&$522.0_{\pm0.0}$ \\
\hline \hline

\end{tabular}%
}
\caption{Baseline for text-video retrieval methods on \datasetName{}.
$\uparrow$ indicates that higher is better (similarly for $\downarrow$).
}
\label{table:retrieval-mnnet} 
\end{table}

%% file: tables/retrieval-modalities.tex
\begin{table}[ht!]
\centering 
\scriptsize 
\setlength{\tabcolsep}{6pt}
\resizebox{\columnwidth}{!}{%
\begin{tabular}{l | @{\hskip 0.1cm}c@{\hskip 0.1cm}c@{\hskip 0.1cm}c@{\hskip 0.1cm}c@{\hskip 0.1cm}c|@{\hskip 0.1cm}c @{\hskip 0.1cm}c@{\hskip 0.1cm}c@{\hskip 0.1cm}c@{\hskip 0.1cm}c@{\hskip 0.1cm}c} 
\hline \hline
\multicolumn{1}{c}{} & 
\multicolumn{5}{c}{Text $\implies$ Video} & \multicolumn{5}{c}{Video $\implies$ Text} \\
Method & R$@$1 $\uparrow$ & R$@$5 $\uparrow$ & R$@$10 $\uparrow$ & MdR $\downarrow$ & MnR $\downarrow$ & R$@$1 $\uparrow$ & R$@$5 $\uparrow$ & R$@$10 $\uparrow$ & MdR $\downarrow$ & MnR $\downarrow$ \\ 
\hline 
SCENE & $8.7_{\pm0.4}$&$26.3_{\pm1.1}$&$37.1_{\pm0.7}$&$22.2_{\pm1.6}$&$52.3_{\pm3.0}$ & $9.1_{\pm0.8}$&$25.4_{\pm0.9}$&$35.3_{\pm1.5}$&$23.2_{\pm0.3}$&$52.6_{\pm2.6}$ \\
PREV. + AUDIO & $7.6_{\pm2.7}$&$27.4_{\pm1.4}$&$40.4_{\pm0.9}$&$17.0_{\pm1.7}$&$49.0_{\pm1.9}$ & $10.1_{\pm1.2}$&$25.7_{\pm1.5}$&$37.5_{\pm1.2}$&$20.0_{\pm1.3}$&$48.9_{\pm2.0}$ \\
PREV. + OBJECTS & $12.7_{\pm1.7}$&$34.8_{\pm1.7}$&$47.0_{\pm1.3}$&$12.3_{\pm0.6}$&$37.6_{\pm2.1}$& $12.8_{\pm1.3}$&$33.5_{\pm2.8}$&$46.6_{\pm1.0}$&$11.8_{\pm0.8}$&$37.6_{\pm1.9}$\\
PREV. + ACTION & $14.3_{\pm0.3}$&$37.5_{\pm1.3}$&$48.6_{\pm0.8}$&$11.3_{\pm0.6}$&$35.2_{\pm1.8}$ & $14.0_{\pm0.3}$&$35.4_{\pm2.9}$&$47.2_{\pm2.8}$&$12.3_{\pm1.5}$&$35.8_{\pm2.4}$ \\

\hline \hline
\end{tabular}%
}
\caption{The influence of different pretrained experts for the performance of the CE model ~\cite{liu2019use} trained on \datasetName{}. The value and cumulative effect of different experts for scene classification (SCENE), ambient sound classification (AUDIO), image classification (OBJECT), and action recognition (ACTION). PREV. denotes the experts used in the previous row.
}
\label{table:ablation-modalities} 
\end{table}

%% file: tables/localization-baselines.tex
\begin{table}
\centering 
\scriptsize 
\setlength{\tabcolsep}{6pt}
\resizebox{\columnwidth}{!}{%
\begin{tabular}{l | @{\hskip 0.1cm}c@{\hskip 0.1cm}c@{\hskip 0.1cm}c@{\hskip 0.1cm}c@{\hskip 0.1cm}c|@{\hskip 0.1cm}c @{\hskip 0.1cm}c@{\hskip 0.1cm}c@{\hskip 0.1cm}c@{\hskip 0.1cm}c@{\hskip 0.1cm}c} 
\hline \hline
\multicolumn{1}{c}{} & 
\multicolumn{5}{c}{Text $\implies$ Video} & \multicolumn{5}{c}{Video $\implies$ Text} \\
Method & R$@$1 $\uparrow$ & R$@$5 $\uparrow$ & R$@$10 $\uparrow$ & MdR $\downarrow$ & MnR $\downarrow$ & R$@$1 $\uparrow$ & R$@$5 $\uparrow$ & R$@$10 $\uparrow$ & MdR $\downarrow$ & MnR $\downarrow$ \\ 
\hline 
E2EWS~\cite{miech2019howto100m} & $6.7_{\pm0.0}$&$14.7_{\pm0.0}$&$20.4_{\pm0.0}$&$133.0_{\pm0.0}$&$342.0_{\pm0.0}$ & $8.4_{\pm0.0}$&$15.4_{\pm0.0}$&$19.8_{\pm0.0}$&$154.5_{\pm0.0}$&$363.0_{\pm0.0}$ \\
MoEE~\cite{miech2018learning} & $19.0_{\pm0.8}$&$38.9_{\pm1.0}$&$47.9_{\pm0.7}$&$12.0_{\pm1.0}$&$127.4_{\pm5.9}$ & $19.8_{\pm0.2}$&$39.6_{\pm0.6}$&$47.6_{\pm0.1}$&$13.0_{\pm0.0}$ & $124.3_{\pm5.5}$\\
CE~\cite{liu2019use} & $18.2_{\pm0.5}$&$38.1_{\pm0.8}$&$46.8_{\pm0.4}$&$13.3_{\pm0.6}$&$127.5_{\pm3.9}$ & $18.1_{\pm0.6}$&$37.3_{\pm0.5}$&$45.9_{\pm0.6}$&$14.0_{\pm1.0}$ & $123.9_{\pm3.3}$\\
\hline \hline
\end{tabular}%
}
\caption{Comparison of localisation methods trained with oracle temporal proposals information on the \datasetName{} dataset.}
\label{table:localization-baselines} 
\end{table}

%% file: conclusion.tex
\section{Conclusion} \label{sec:conclusion}

In this work, we introduced the \datasetName{} dataset for video understanding. 
Through extensive analysis, we demonstrated that this dataset contains detailed annotations with rich, highly-relevant vocabulary.
To demonstrate its applications and to bootstrap the development of future methods, we evaluated several baselines for video retrieval and clip localisation with an oracle.
By directing researchers' attention to this video-to-audio narration problem, and providing them with the tools to tackle it,
we hope that \emph{YouDescribe}'s goals of assisting the visually-impaired will be fully realised.
\vspace{4pt}

%% file: ack.tex
\paragraph{Acknowledgements.}
{
This work is supported by the EPSRC (VisualAI EP/T028572/1 and DTA Studentship), and the Royal Academy of Engineering (DFR05420).
We are grateful to Sophia Koepke for her helpful comments and suggestions.
}